\documentclass{article}

    \usepackage[preprint,nonatbib]{neurips_2024}

\usepackage[utf8]{inputenc} 
\usepackage[T1]{fontenc}    
\usepackage{hyperref}       
\usepackage{url}            
\usepackage{booktabs}       
\usepackage{amsfonts}       
\usepackage{nicefrac}       
\usepackage{microtype}      
\usepackage{xcolor}         
\usepackage{graphicx}

\title{Enhancing Robustness of Human Detection Algorithms in Maritime SAR through Augmented Aerial Images to Simulate Weather Conditions}

\author{
  Miguel Tjia\thanks{Lead Author}\\
  Jakarta Intercultural School\\
  Jakarta Selatan, DKI Jakarta, Indonesia \\
  \texttt{migueltjia88@gmail.com} \\
  \And
  Artem Kim$^{*}$\\
  Busan Foreign School\\
  Haeundae, Busan, South Korea  \\
  \texttt{2025kimat@bfs.or.kr} \\
  \And
  Elaine Wynette Wijaya \\
  SMPK 8 Penabur Tanjung Duren\\
  Jakarta Barat, DKI Jakarta, Indonesia \\
  \texttt{elainewijaya97@gmail.com} \\
  \And
  Hanna Tefera\\
  Avon High School\\
  Avon, Indiana, United States of America  \\
  \texttt{hannatefera16@gmail.com} \\
  \And
  Kevin Zhu\thanks{Senior Author}\\
  Algoverse AI Research\\
  Palo Alto, CA, United States of America  \\
  \texttt{kevin@algoverse.us} \\
}

\begin{document}

\maketitle
\begin{abstract}
7,651 cases of Search and Rescue Missions (SAR) were reported by the United States Coast Guard in 2024, with over 1322 SAR helicopters deployed in the 6 first months alone \cite{US}. Through the utilizations of YOLO, we were able to run different weather conditions and lighting from our augmented dataset for training. YOLO then utilizes CNNs to apply a series of convolutions and pooling layers to the input image, where the convolution layers are able to extract the main features of the image \cite{CNN}. Through this, our YOLO model is able to learn to differentiate different objects which may considerably improve its accuracy, possibly enhancing the efficiency of SAR operations through enhanced detection accuracy. This paper aims to improve the model's accuracy of human detection in maritime SAR by evaluating a robust datasets containing various elevations and geological locations, as well as through data augmentation which simulates different weather and lighting. We observed that models trained on augmented datasets outperformed their non-augmented counterparts in which the human recall scores ranged from 0.891 to 0.911 with an improvement rate of 3.4\% on the YOLOv5l model. Results showed that these models demonstrate greater robustness to real-world conditions in varying of weather, brightness, tint, and contrast.
\end{abstract}

\section{Introduction}
The growing number of commercial vessels and pleasure craft in these convoluted waters, coupled with the complex coastal environment, highlights the need for a comprehensive and effective SAR (Search and Rescue) services in case a maritime incident or accident occurs \cite{clearseas-msr-canada}. From the 1st of January to the 16th of June of 2024, the United States alone reported over 7,651 cases of Search and Rescue Missions, having saved 2,430 lives \cite{US}. Furthermore, traditional rescue methods often face limitations where the weather, presence of people \& ships and drift situations may lead to delays in doing a rescue mission \cite{imo-large-scale-rescue}. Since survival rates drop exponentially during the first 18 hours \cite{adams}, finding missing people may prove to be an operation's most important and challenging aspects. Hence why SAR operations may be revolutionized by using deep learning to process images taken from UAVs (unnamed aerial vehicles).

 Therefore, our aim is to contribute to this problem by creating a robust and efficient benchmark that can accurately identify and differentiate humans from different classes of boats, kayaks and buoys. \\

\textbf{Our contributions are as follows:}
\begin{enumerate}
    \item \textbf{Robust Dataset:} We developed a comprehensive way of creating a robust dataset across varying environmental conditions. Through the process of data augmentation, we were able to simulate different geological locations, weather conditions and lighting.
    \item \textbf{Improving Robustness of YOLO models:} Using different models of YOLO, we aim to highlight the difference in the performance of YOLO models when trained with augmented dataset and a non-augmented dataset.
\end{enumerate}

\section{Related Works}

Deep learning algorithms such as YOLO models, have shown extensive promise in the field of SAR efforts up until now. As YOLO are advanced at analyzing visual data, they are suitable for monitoring activities in aquatic environments and have displayed effectiveness in detecting objects in the water using aerial images \cite{sharafaldeen2022marine}.

Bachir and Memon \cite{BACHIR2024100243}  evaluated the performance of YOLOv5l in detecting humans on land for search and rescue operations. In the paper, the author found that YOLOv5l performed exceptionally well in identifying humans under clear sunlight conditions, achieving high recall and precision rates. However, its performance deteriorated when humans were in shadowed areas, where recall dropped from 0.932 to 0.893 after using a dataset mixed with shadow-containing images. This observation underscores the sensitivity of YOLOv5l to lighting and weather conditions, suggesting its lack of robustness in challenging environments commonly encountered in real-life SAR scenarios. 

\section{Methodology}\label{headings}

\subsection{Datasets}
In this study, two publicly available datasets comprising aerial images of various aquatic zones, including coastal areas, lakes, and oceans, were merged. All images from both datasets were captured using drone-mounted cameras with high elevations from a bird's-eye view. The first dataset, AFO (Aerial dataset of Floating Objects) \cite{article}, contains images of various floating objects commonly found in aquatic environments, such as humans and inanimate objects including boats, surfboard, sailboats, and kayaks. Meanwhile, the second dataset \cite{roger-water_dataset} captures images of human swimming and floating. The combined dataset includes 13,282 raw aerial images, each with a compressed resolution of 640 pixels. 

\subsubsection{Data Augmentation}
The merged dataset initially contains only daylight images captured under clear weather conditions. However, to ensure the model is applicable for real-life conditions where weather can be unpredictable, we artificially simulated various weather conditions in different times of day, including foggy, rainy, and sunny conditions, with the latter having more brightness compared to clear weather. We created duplicates of the aerial images and applied several synthetic transformations to their attributes. Specifically, the fog effect was achieved by blending the images with a white layer, sunny conditions were simulated by increasing brightness and applying a yellow tint, rain was emulated by overlaying a rain texture, and nighttime conditions were created by reducing brightness and contrast and adding a blue tint. From this augmentation, the dataset was expanded to have a split of 50\% day-clear, while the remaining is split evenly through six different categories of weather: day-rain, day-cloudy, day-sunny, nighttime-clear, night-rain, and night-cloudy. As a result of the data augmentation, our final dataset comprises a total of 26,548 images. Figure 1 shows the sample of our dataset augmentation. 

\begin{figure}[ht]
    \centering
    \includegraphics[width=14cm, height = 4cm]{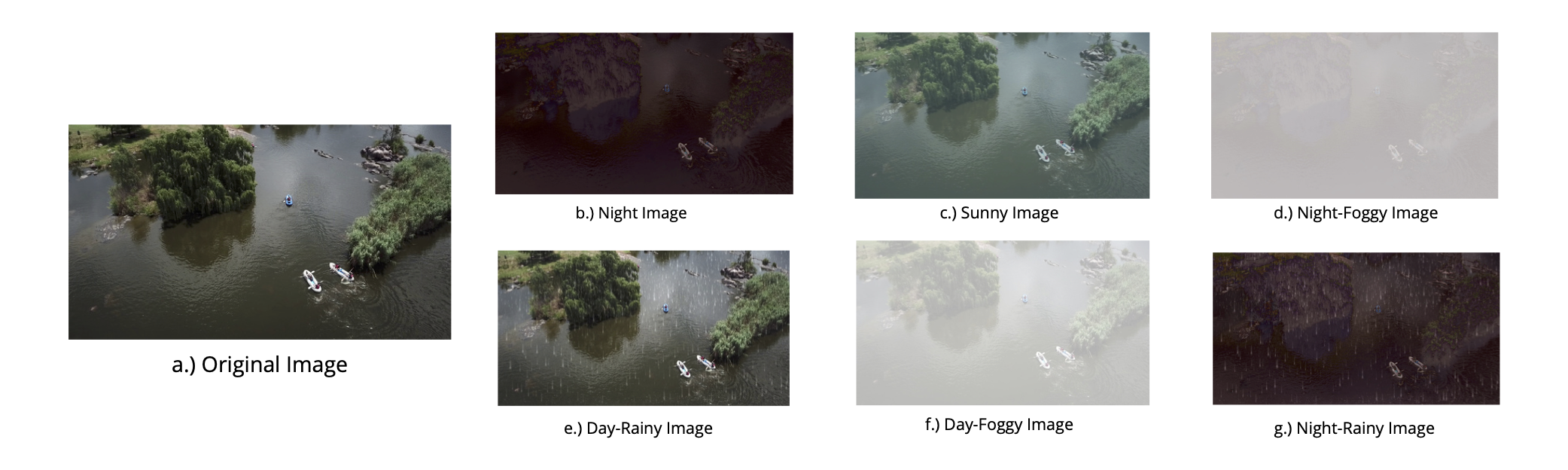}
    \caption{Sample Dataset Augmentation}
    \label{fig:sample-label}
\end{figure}

\subsection{Method}

\subsubsection{YOLO \& Training}
In this study, we utilized four different algorithms for training:  YOLOv5s, YOLOv5l, YOLOv10s, and YOLOv10l. The basic framework of all YOLO algorithms uses bounding boxes, CNNs (convolutional neural network), and class probabilities to make predictions all at once \cite{kundu2023yolo}, while YOLOv10 utilizes non-maximum suppression (NMS) to remove duplicate bounding boxes \cite{viso-yolov10}. Furthermore, the small and large models have a trade-off between speed and accuracy respectively. The procedure begins with acquiring a robust dataset, containing augmented images of different weather conditions, geological locations, elevations and lighting. The images were then split into training, validation, and testing sets in a 70-15-15 ratio, to ensure robust model evaluation. Through the training process, we ensured that each synthetic augmentation only showed up once, in either training, validation or testing sets. This allowed us to successfully split the data without data leakage which may lead to unintended bias. We then trained the model using: 52 epochs and 32 batches per training session with a learning rate of 0.01. 

\subsubsection{Evaluation Metrics}
We are able to achieve different evaluation metrics on our model’s results i.e. Precision, Recall, F1, mAP@0.5, mAP@0.5-0.95 which are then used to assess the performance of the model. The performance of both humans and inanimate objects were highly assessed to ensure the model’s applicability in detecting humans in different environments, as without different object classes or images with no humans, the model is more prone to falsely classify inanimate objects as human. Even so, our research optimizes recall accuracy through our training process, since the consequences of misclassifying a human as an inanimate object, such as a boat, potentially resulting in a loss of life is much more severe than mistaking an inanimate object as human. 

\section{Results and Discussions}
The performance of our YOLO models(YOLOv5l, YOLOv5s, YOLOv10l, YOLOv10s) trained with augmented dataset models is evaluated using precision, recall, F1 score, and mean Average Precision (mAP). Table \ref{perfomance-table} presents the evaluation metrics for each model. 

Generally, our YOLO models are effective in detecting human objects, with an overall human recall ranging from 0.86 to 0.91, with YOLOv5l emerging as the top-performing model among those tested and YOLOv5s as the bottom-performing model. Moreover, the YOLO models exhibit a great balance between recall and precision, meaning high accuracy and completeness in identifying humans. YOLOv5l shows the best balance with an F1 score of 0.94, while YOLOv5s has the lowest balance with an F1 score of 0.89. On the other hand, the YOLO models also perform well in detecting inanimate objects, with recall rates ranging from 0.85 to 0.95. 

When compared with the baseline YOLO models (trained without dataset augmentation), our YOLOv10l-AD exhibits clear superiority over its baseline counterpart, increasing human recall rates from 0.87 to 0.91, as shown in Figure 2. Moreover, when compared to another study that trains YOLOv5l on the non-augmented AFO dataset \cite{mahmoud2024enhancing}, our YOLOv5l shows superior performance in detecting human objects compared to its baseline version, with recall rates improving from 0.89 to 0.91. One possible reason for this improvement is that the augmented dataset enhances the models' robustness to variations in image properties such as brightness, tint, and contrast, potentially explaining their improved overall performance in tackling unfamiliar conditions.

\begin{table}[ht]
  \caption{Performance of YOLO models trained with augmented dataset and YOLOv10l without Augmented Dataset}
  \label{perfomance-table}
  \centering
  \begin{tabular}{lllllll}
    \toprule
    Algorithm & Class & Precision & Recall & F1 & mAP50 & mAP50-90 \\
    \midrule
    YOLOv5s & All & 0.93  & 0.91 & 0.92 & 0.93 & 0.60  \\
          (augmented dataset)  & Humans & 0.94  & 0.86 & 0.89 & 0.92 & 0.52  \\
            & Inanimate objects & 0.93  & 0.92 & 0.93 & 0.94 & 0.61  \\
    YOLOv5l & All & \textbf{0.95}  & \textbf{0.94} & \textbf{0.95} & \textbf{0.95} & \textbf{0.67}  \\
           (augmented dataset) & Humans & \textbf{0.96} & \textbf{0.91} & \textbf{0.94} & \textbf{0.96} & \textbf{0.63}      \\
            & Inanimate objects & \textbf{0.95} & \textbf{0.95} & \textbf{0.95} & \textbf{0.95} & \textbf{0.68}      \\
    YOLOv10s & All & 0.92  & 0.85 & 0.88 & 0.91 & 0.62  \\
            (augmented dataset) & Humans & 0.94  & 0.86 & 0.9 & 0.9 & 0.6  \\
             & Inanimate objects & 0.92  & 0.85 & 0.86 & 0.90 & 0.63  \\
    YOLOv10l & All & 0.94  & 0.89 & 0.91 & 0.92 & 0.66  \\
             (augmented dataset) & Humans & 0.93 & \textbf{0.91} & 0.92 & 0.95 & \textbf{0.63}      \\
             & Inanimate objects & 0.94 & 0.89 & 0.91 & 0.92 & 0.67      \\
    YOLOv10l & All & 0.91 & 0.85 & 0.88 & 0.91 & 0.61  \\
                               (non-augmented dataset)& Humans & 0.94 & 0.87 & 0.90 & 0.93 & 0.56      \\
                               & Inanimate objects & 0.90 & 0.85 & 0.86 & 0.90 & 0.61     \\
    \bottomrule
  \end{tabular}
\end{table}

\begin{figure}[ht]
    \centering
    \includegraphics[width=10cm]{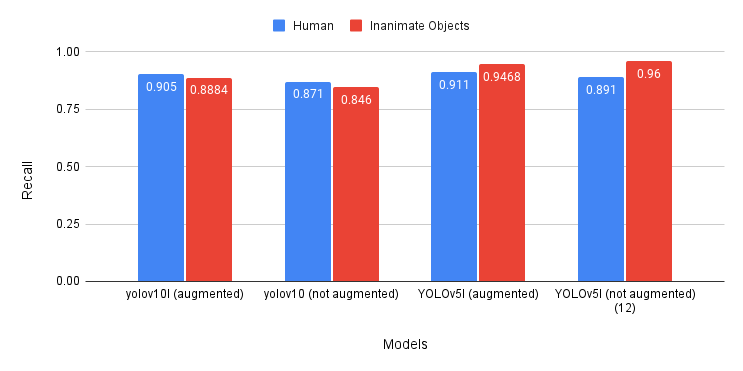}
    \caption{Comparison of YOLOv5l and YOLOv10l trained with Augmented Dataset With Their Baseline (Not Trained with Augmented Dataset) Counterpart}
    \label{fig:enter-label}
\end{figure}

\section{Conclusion and Future Works}

This study aimed to improve the performance of YOLO models for maritime SAR operations by augmenting aerial images to simulate various weather conditions and lighting as well as combining different datasets with various geographical and elevations features. Our findings illustrate that the YOLO models trained with augmented datasets excel in comparison to baseline YOLO models that were trained with non-augmented datasets. Specifically, for both YOLOv5l and YOLOv10l, utilizing augmented datasets resulted in an increase in human object recall. Our YOLOv5l-Augmented Dataset achieved a recall of 0.91, compared to its baseline version at 0.89, while our YOLOv10l-Augmented Dataset achieved a recall of 0.91, compared to its baseline at 0.87.

Future research could explore integrating advanced generative models such as CycleGAN to enhance the realism of augmented aerial images depicting simulated weather conditions. These models are adept at learning mappings across diverse weather scenarios, generating images that exhibit nuanced variations and naturalistic. \cite{weatherGAN} By leveraging these techniques, our goal is to enhance the model's ability to maintain robust performance in unpredictable environmental settings.

\newpage
\section{Acknowledgments}
Conceptualization: M. Tjia, A. Kim \\[0.4 em]
Data Curating: A. Kim, M. Tjia, E.W. Wijaya \\[0.4 em]
Software \& Coding: A. Kim \\[0.4 em]
Training: A. Kim, E.W. Wijaya \\[0.4 em]
Writing - Abstract: E.W. Wijaya\\[0.4 em]
Writing - Introduction: E.W. Wijaya\\[0.4 em]
Writing - Related Works: M. Tjia, H. Tefera\\[0.4 em]
Writing - Methodology: M. Tjia, E.W. Wijaya\\[0.4 em]
Writing - Results and Discussions: M. Tjia\\[0.4 em]
Writing - Conclusion: M. Tjia\\[0.4 em]
Writing - Review \& Editing: M. Tjia, E.W. Wijaya, A. Kim\\[0.4 em]
Visualization: A. Kim, M. Tjia, E.W. Wijaya\\[0.4 em]
Supervisions: Kevin Zhu gave lectures on ML and research skills, suggested readings and high-level guidance, and gave comments on paper.

\medskip
\newpage

\small
\bibliographystyle{ieeetr}
\bibliography{main}

\end{document}